\documentclass[letterpaper, 10 pt, conference]{ieeeconf}  

\IEEEoverridecommandlockouts                              

\usepackage{graphics} 
\usepackage{epsfig} 
\usepackage{mathptmx} 
\usepackage{amsmath} 
\usepackage{amssymb}  
\usepackage{lipsum}
\usepackage{bm}
\usepackage{xcolor}
\usepackage{comment}

\usepackage[style=ieee]{biblatex}

\DeclareSourcemap{
  \maps{
    \map{
      \pertype{article}
      \step[fieldset=language, null]
      \step[fieldset=url, null]
      \step[fieldset=doi, null]
      \step[fieldset=issn, null]
      \step[fieldset=isbn, null]
      \step[fieldset=note, null]
      \step[fieldset=editor, null]
      \step[fieldset=urldate, null]
      \step[fieldset=file, null]
    }
  }
}
\DeclareSourcemap{
  \maps{
    \map{
      \pertype{inproceedings}
      \step[fieldset=language, null]
      \step[fieldset=url, null]
      \step[fieldset=doi, null]
      \step[fieldset=issn, null]
      \step[fieldset=isbn, null]
      \step[fieldset=note, null]
      \step[fieldset=editor, null]
      \step[fieldset=urldate, null]
      \step[fieldset=file, null]
    }
  }
}
\DeclareSourcemap{
  \maps{
    \map{
      \pertype{incollection}
      \step[fieldset=language, null]
      \step[fieldset=url, null]
      \step[fieldset=doi, null]
      \step[fieldset=issn, null]
      \step[fieldset=isbn, null]
      \step[fieldset=note, null]
      \step[fieldset=editor, null]
      \step[fieldset=urldate, null]
      \step[fieldset=file, null]
    }
  }
}

\title{\LARGE \bf
Quadratic Programming Optimization for Bio-Inspired Thruster-Assisted Bipedal Locomotion on Inclined Slopes
}

\author{Shreyansh Pitroda$^{1}$, Eric Sihite$^{2}$, Kaushik Venkatesh Krishnamurthy$^{1}$, \\ Chenghao Wang$^{1}$, Adarsh Salagame$^{1}$, Reza Nemovi$^{2}$, Alireza Ramezani$^{1*}$, and Morteza Gharib$^{2}$
\thanks{$^{1}$ The authors are with SiliconeSynapse Labs, the Department of Electrical Engineering, Northeastern University, USA.}%
\thanks{$^{2}$ The authors are with the Department of Aerospace Engineering, California Institute of Technology, USA.}%
\thanks{$^{*}$ The corresponding author. Email: a.ramezani@northeastern.edu}%
}

\begin{document}

\maketitle
\thispagestyle{empty}
\pagestyle{empty}


\begin{abstract}

Our work aims to make significant strides in understanding unexplored locomotion control paradigms based on the integration of posture manipulation and thrust vectoring. These techniques are commonly seen in nature, such as Chukar birds using their wings to run on a nearly vertical wall. In this work, we show quadratic programming with contact constraints which is then given to the whole body controller to map on robot states to produce a thruster-assisted slope walking controller for our state-of-the-art Harpy platform. Harpy is a bipedal robot capable of legged-aerial locomotion using its legs and thrusters attached to its main frame. The optimization-based walking controller has been used for dynamic locomotion such as slope walking, but the addition of thrusters to perform inclined slope walking has not been extensively explored. In this work, we derive a thruster-assisted bipedal walking with the quadratic programming (QP) controller and implement it in simulation to study its performance.

\end{abstract}

\color{black}
\section{Introduction}

Legged robots offer a significant advantage over wheeled robots due to their ability to navigate complex and unstructured environments like forests, obstacles, and debris. However, controlling legged robots presents intricate challenges related to the hybrid nature of forces needed for movement, as the robot must establish and break contact with the ground using its feet. Initially, control design for legged robots relied on heuristic approaches, which achieved notable successes, such as Raibert’s \cite{raibert_dynamically_nodate} walking controller, Jerry Pratt's \cite{pratt_capture_2006-1} capture point control, and Focchi's \cite{focchi_heuristic_2020} heuristic locomotion planning for quadrupedal robots. Despite their success, heuristic methods have several limitations: 1) they are not easily generalizable to various terrains and motions; 2) they cannot predict the robot’s future state, thus failing to ensure the physical feasibility of planned trajectories. These limitations have driven research toward optimization-based predictive locomotion planning to avoid the short-sighted behaviors inherent in heuristic methods.

There are many recent work that use optimization techniques to traverse inclined slopes \cite{ding_real-time_2019,kim_computationally-robust_2018,he_mechanism_2020}.  By framing locomotion planning as an optimization problem, high-level locomotion tasks can be represented as cost functions, and system dynamics can be incorporated as constraints. In addition to robot dynamics, locomotion tasks must also adhere to contact dynamics, such as unilateral force and friction cone constraints, which are crucial for stabilizing locomotion.

\begin{figure}[t]
    \vspace{0.05in}
    \centering
    \includegraphics[width=1\linewidth]{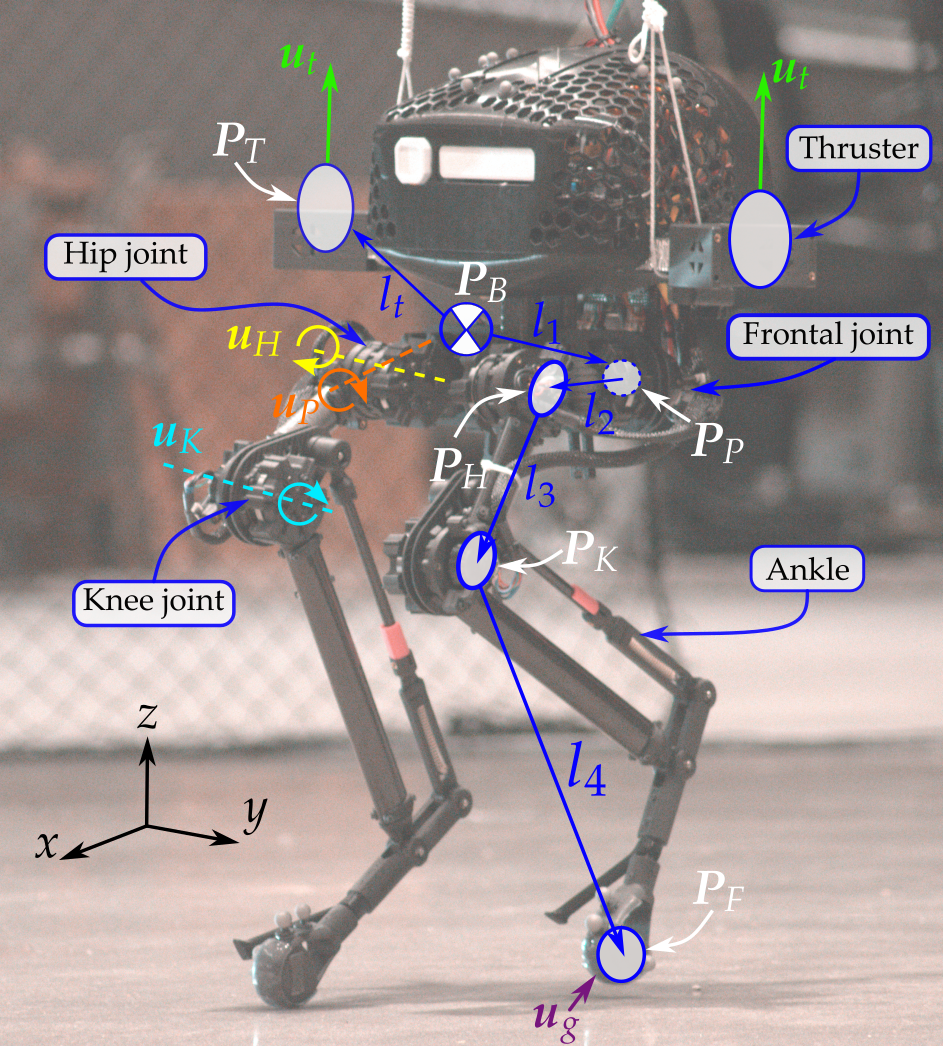}
    \caption{Illustrates the Harpy platform, a bipedal robot with a thruster attached on each side which motivates our wing-assisted inclined walking.}
    \label{fig:cover-image}
    \vspace{-0.05in}
\end{figure}

The primary contribution of this work is the development of a quadratic programming (QP) solver with contact and dynamic constraints. This combination is used to create a thruster-assisted walking controller on $30^{\circ}$ Inclined slope for our state-of-the-art Harpy platform, as illustrated in Fig.~\ref{fig:cover-image}. 

Recent tests have explored the application of thrusters (i.e., thrust vectoring) and posture manipulation in notable robots such as the Multi-modal mobility morphobot (M4) \cite{sihite_multi-modal_2023, sihite_efficient_2022, mandralis_minimum_2023} and LEONARDO \cite{kim_bipedal_2021, liang_rough-terrain_2021, sihite_optimization-free_2021, sihite_efficient_2022,pitroda_capture_2024-1}. 

The M4 robot aims to enhance locomotion versatility by combining posture manipulation and thrust vectoring, enabling modes such as walking, wheeling, flying, and loco-manipulation. LEONARDO, a quadcopter with two legs, can perform both quasi-static walking and flying. However, neither of these robots fully demonstrate dynamic legged locomotion and aerial mobility. Integrating these modes poses a significant challenge due to conflicting requirements (see \cite{sihite_multi-modal_2023}).

Our work aims to make significant strides in understanding unexplored locomotion control paradigms based on the integration of posture manipulation and thrust vectoring. These techniques are commonly used by birds which are known for their locomotion plasticity and robust locomotion feats. For instance, Chukar birds can perform a wing-assisted incline running (WAIR) maneuver \cite{dial_wing-assisted_2003-1, tobalske_aerodynamics_2007}. In the WAIR maneuver, Chukar birds utilize their flapping wings and the resulting aerodynamic forces to increase contact forces, allowing them to ascend steep slopes that conventional bipedal robots would find challenging to navigate.

In this study, we employ a detailed model of Harpy (depicted in Fig.~\ref{fig:cover-image}) using Matlab Simscape to evaluate our controller's effectiveness. Harpy is equipped with eight custom-designed high-energy density actuators for dynamic walking, along with electric ducted fans mounted on its torso sides. Harpy's height measures 600 cm and weighs 6 Kg. It hosts a computer based on Elmo amplifiers for real-time low-level control command executions.

Harpy's design integrates advantages from both aerial and dynamic bipedal-legged systems. Currently, the hardware design and assembly of Harpy have been completed~\cite{pitroda_dynamic_2023}, and our primary goal is to explore various control design strategies for this platform \cite{dangol_control_2021,pitroda_capture_2024,ramezani_performance_2014,Ramezani2013FeedbackCD}.

In this work, we aim to design a planning and control approach that enables stable walking on an inclined slope. The main contributions of this paper are: a) formulating a QP-based reference tracking controller which has dynamics of VLIP model. b) Whole body mapping which maps the output of QP solver to whole body dynamics.

This work is structured as follows: we present the derivations of the Harpy high fidelity model, the reduced order model, followed by the Swing leg controller, Online planning using Quadratic Programming(QP), Whole body mapping, simulation results, and concluding remarks.

\section{Thruster-Assisted Bipedal Walking Model}
\begin{figure}[t]
    \vspace{0.05in}
    \centering
    \includegraphics[width=1\linewidth]{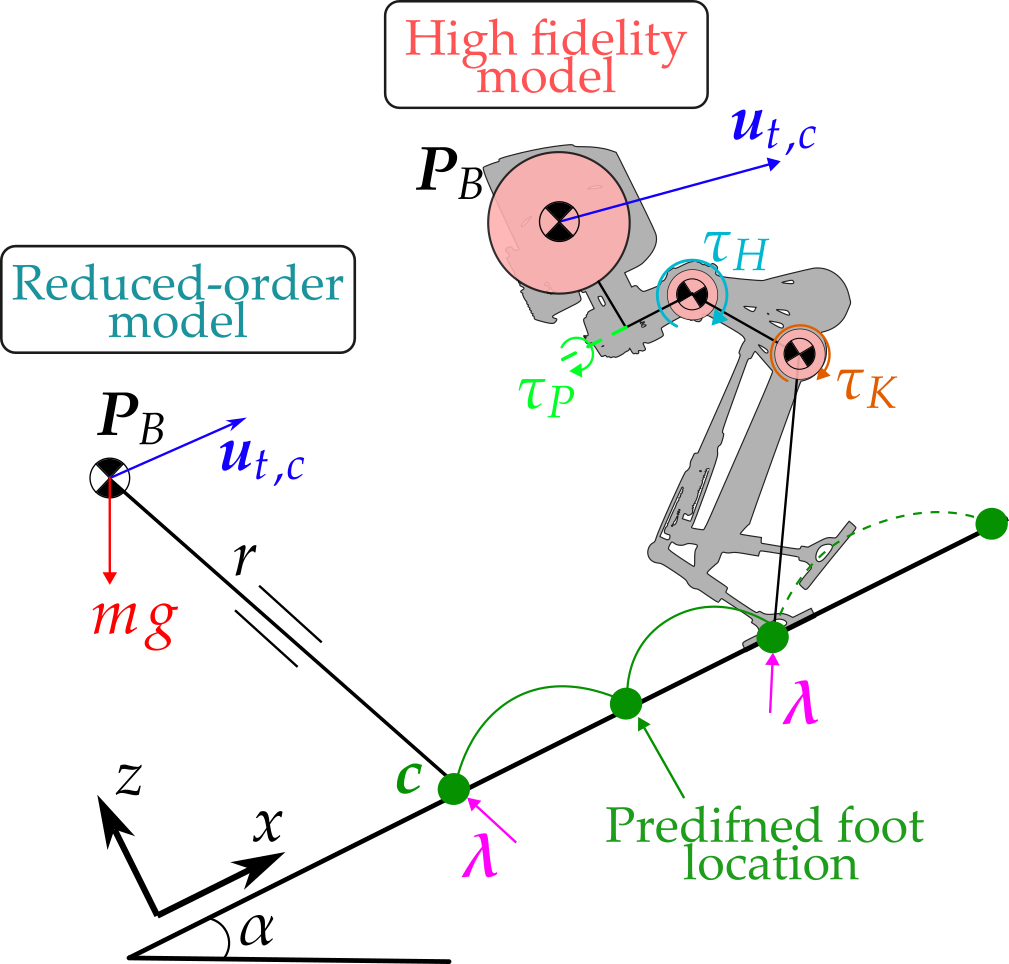}
    \caption{Illustrates Reduced-order model (ROM) and High fidelity model (HFM). ROM is represented as a point mass and massless link. HFM is represented as three inertia bodies which is used in numerical simulation.}
    \label{fig:rom-model}
    \vspace{-0.05in}
\end{figure}
This section outlines the dynamics formulation of the robot which is used in the numerical simulation, in addition to the reduced order models which are used in the controller design. Figure \ref{fig:cover-image} shows the kinematic configuration of Harpy which lists the center of mass (CoM) positions of the dynamic components, joint actuation torques, and thruster torques. The system model has a combined total of 12 degrees-of-freedoms (DoFs): 6 for the body and 3 for each leg. Due to the symmetry, the left and right sides of the robot follow similar derivations so only the general derivations are provided in this section.

\subsection{Energy-based Lagrange Formalism}
The Harpy equations of motion are derived using the Euler-Lagrangian dynamics formulation. In order to simplify the system, each link is assumed to be mass-less, with the mass and inertia concentrated at the body and each motor. Consequently, the lower leg kinematic chain is considered massless, significantly simplifying the system. The three leg joints are labeled as the hip frontal (pelvis $P$), hip sagittal (hip $H$), and knee sagittal (knee $K$), as illustrated in Fig. \ref{fig:cover-image}. The thrusters are also considered massless and capable of providing forces in any direction to simplify the problem.

Let $\gamma_h$ be the frontal hip angle, while $\phi_h$ and $\phi_k$ represent the sagittal hip and knee angles, respectively. The superscripts $\{B,P,H,K\}$ represent the frame of reference about the body, pelvis, hip, and knee, while the inertial frame is represented without the superscript. Let $R_B$ be the rotation matrix from the body frame to the inertial frame (i.e., $\bm x = R_B\, \bm x^B$). The pelvis motor mass is added to the body mass. Then, the positions of the hip and knee centers of mass (CoM) are defined using kinematic equations:
\begin{equation}
\begin{gathered}
    \bm{p}_P = \bm{p}_{B} + R_{B}\, \bm{l}_{1}^{B}, \\
    \bm{p}_H = \bm{p}_{P} + R_{B}\,R_x(\gamma_h)\, \bm{l}_{2}^{P} \\
    \bm{p}_K = \bm{p}_{H} + R_{B}\,R_x(\gamma_h)\,R_y(\phi_h) \bm{l}_{3}^{H},
\end{gathered}
\label{eq:pos_com}
\end{equation}
where $R_x$ and $R_y$ are the rotation matrices about the $x$ and $y$ axes, respectively, and $\bm l$ is the length vector representing the configuration of Harpy, which remains constant in its respective local frame of reference. The positions of the foot and thrusters are defined as:
\begin{equation}
\begin{gathered}
    \bm{p}_F = \bm{p}_{K} + R_{B}\,R_x(\gamma_h)\,R_y(\phi_h)\,R_y(\phi_k)\, \bm{l}_{4}^{K} \\
    \bm{p}_T = \bm{p}_{B} + R_{B}\, \bm{l}_{t}^{B}
\end{gathered}
\label{eq:pos_other}
\end{equation}
where the length vector from the knee to the foot is $\bm l_4^K = [-l_{4a}\cos{\phi_k}, 0, -( l_{4b} + l_{4a}\sin{\phi_k})]^\top$, which represents the kinematic solution to the parallel linkage mechanism of the lower leg. Let $\bm \omega_B$ be the angular velocity of the body. Then, the angular velocities of the hip and knee are defined as: 
\begin{equation}
  \begin{aligned}  
        \bm \omega_H^B &= [\dot{\gamma}_h,0,0]^\top + \bm \omega_B^B\\
        \bm \omega_K^H &= [0,\dot{\phi}_h,0]^\top + \bm \omega_H^H 
    \end{aligned}
\end{equation}
Consequently, the total energy of Harpy for the Lagrangian dynamics formulation is defined as follows:
\begin{equation}
\begin{aligned}
    K &= \tfrac{1}{2} \textstyle \sum_{i \in \mathcal{F}} \left( 
        m_i\,\bm p_i^\top\, \bm p_i + 
        \bm \omega_i^{i \top} \, \hat I_i \, \bm \omega_i^i \right) \\
    V &= - \textstyle \sum_{i \in \mathcal{F}} \left( 
        m_i\,\bm p_i^\top\, [0,0,-g]^\top \right),
\end{aligned}
\label{eq:energy}
\end{equation}
\begin{figure}[t]
    \vspace{0.05in}
    \centering
    \includegraphics[width=0.9\linewidth]{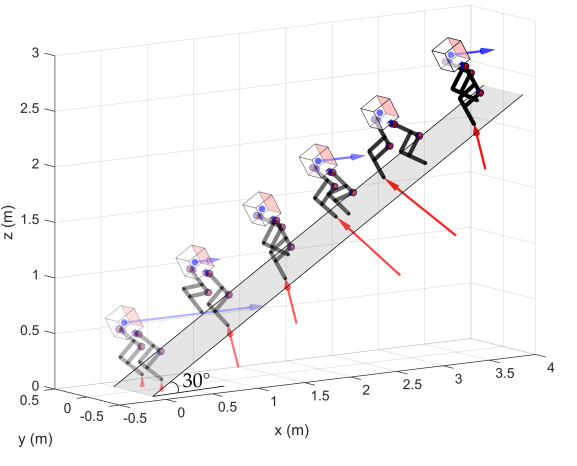}
    \caption{Snapshot of the simulation result, showing thruster-assisted inclined slope walking using the QP controller.}
    \label{fig:snapshot}
    \vspace{-0.05in}
\end{figure}
In this context, $\mathcal{F} = \{B,H_L,K_L,H_R,K_R\}$ denotes the relevant frames of reference and mass components, including the body, left hip, left knee, right hip, right knee. The subscripts L and R indicate the left and right sides of the robot, respectively. Additionally, $\hat I_i$ represents the inertia relative to its local frame, and $g$ is the gravitational constant. This forms the Lagrangian of the system, expressed as $L = K - V$, which is used to derive the Euler-Lagrange equations of motion. The body's angular velocity dynamics are obtained using a modified Lagrangian for rotation in $SO(3)$, avoiding the use of Euler angles and the associated risk of gimbal lock. Consequently, the following equations of motion are based on Hamilton's principle of least action:
\begin{equation}
\begin{gathered}
    \tfrac{d}{dt}\left( \tfrac{\partial L}{\partial \bm \omega_B^B}  \right) + 
    \bm \omega_B^B \times \tfrac{\partial L}{\partial \bm \omega_B^B} + 
    \textstyle \sum_{j=1}^{3} \bm r_{Bj} \times \tfrac{\partial L}{\partial \bm r_{Bj}} = \bm u_1, \\
    \tfrac{d}{dt}\left( \tfrac{\partial L}{\partial \dot {\bm q}}  \right) - 
    \tfrac{\partial L}{\partial \bm q} = \bm u_2, \\ 
    \tfrac{d}{dt} R_B = R_B\, [\bm \omega_B^B]_\times,
\end{gathered}
\label{eq:eom_eulerlagrange}
\end{equation}
where $[\, \cdot \, ]_\times$ denotes the skew symmetric matrix, $R_B^\top = [\bm r_{B1}, \bm r_{B2}, \bm r_{B3}]$, $\bm q = [\bm p_B^\top, \gamma_{h_L}, \gamma_{h_R}, \phi_{h_L}, \phi_{h_R}]^\top$ represents the dynamical system states other than $(R_B,\bm \omega^B_B)$, and $\bm u$ denotes the generalized forces. There is no mass associated with the knee sagittal angle $\phi_k$, and thus, it is updated using the joint acceleration input $\bm u_k = [\ddot{\phi}_{k_L}, \ddot{\phi}_{k_R}]^\top$. Then, the system acceleration can be derived as follows: 
\begin{equation}
\begin{gathered}
    M \bm a + \bm h = B_j\, \bm u_j + \bm u_t + B_g\, \bm u_g
\end{gathered}
\label{eq:eom_accel}
\end{equation}
where $\bm a = [ \dot{\bm \omega}_B^{B\top}, \ddot{\bm q}^\top, \ddot{\phi}_{k_L}, \ddot{\phi}_{k_R}]^\top$, $\bm u_t$ denotes the generalized thruster force, $\bm u_j = [u_{P_L}, u_{P_R}, u_{H_L}, u_{H_R}, \bm u_k^\top]^\top$ represents the joint actuation, and $\bm u_g$ stands for the ground reaction forces (GRFs). The variables $M$, $\bm h$ and $B_g$ are functions of the full system states:
\begin{equation}
    \bm x = [\bm r_{B}^\top, \bm q^\top, \phi_{K_L}, \phi_{K_R}, \bm \omega_B^{B \top}, \dot{\bm q}^\top, \dot{\phi}_{K_L}, \dot{\phi}_{K_R}]^\top,
\label{eq:states}
\end{equation}
where the vector $\bm r_B$ contains the elements of $R_B$. Introducing $B_j = [0_{6 \times 6}, I_{6 \times 6}]$ allows $\bm u_j$ to actuate the joint angles directly. Let $\bm v = [\bm \omega_B^{B\top}, \dot{\bm q}^\top]^\top$ denote the velocity of the generalized coordinates. Then, $B_t$ and $B_g$ can be defined using the virtual displacement from the velocity as follows:
\begin{equation}
\begin{aligned}
    B_g = \begin{bmatrix}
        \begin{pmatrix}
        \partial \dot{\bm p}_{F_L} / \partial \bm v \\
        \partial \dot{\bm p}_{F_R} / \partial \bm v
        \end{pmatrix}^\top
        \\
        0_{2 \times 6}
    \end{bmatrix}.
\end{aligned}
\label{eq:generalized_forces}
\end{equation}
%
%
\subsection{Compliant Ground Model}

The GRF is modeled using the unilateral compliant ground model with undamped rebound, while friction is modeled using the Stribeck friction model, defined as follows:
\begin{equation}
\begin{aligned}
    u_{g,z} =& -k_{g,p}\, p_{F,z} - k_{g,d}\, \dot p_{F,z} \\
    u_{g,x} =& -\left(\mu_c + (\mu_s - \mu_c)\, \mathrm{exp}\left(-\tfrac{|\dot p_{F,x}|^2}{v_s^2}\right) \right) f_z\, \mathrm{sgn}(\dot p_{F,x}) \\ 
    & - \mu_v\,\dot p_{F,x},
\end{aligned}
\label{eq:ground_model}
\end{equation}
where $p_{F,x}$ and $p_{F,z}$ represent the $x$ and $z$ components of the inertial foot position, $k_{g,p}$ and $k_{g,d}$ denote the spring and damping model for the ground, $\mu_c$, $\mu_s$, and $\mu_v$ are the Coulomb, static, and viscous friction coefficients, respectively, and $v_s$ is the Stribeck velocity. $k_{g,d}$ is set to $0$ if $\dot p_{F,z} > 0$ for the undamped rebound model, and friction in the $y$ direction follows a similar derivation to $u_{g,x}$. Then, the ground force model $\bm u_g$ is defined as follows:
\begin{equation}
\begin{aligned}
    \bm u_g = [\bm u_{g_L}^\top\, H(-p_{F_L,z}),\, \bm u_{g_R}^\top\, H(-p_{F_R,z})]^\top,
\end{aligned}
\label{eq:ground_forces}
\end{equation}
where $H(x)$ denotes the Heaviside function, while $\bm u_{g_L}$ and $\bm u_{g_R}$ represent the left and right ground forces, which are formed using their respective components $u_{g,x}$, $u_{g,y}$, and $u_{g,z}$. 

The full-dynamics model can be derived using equations \eqref{eq:eom_eulerlagrange} to \eqref{eq:ground_forces} to form 
\begin{equation}
    \dot{\bm x} = \bm f(\bm x, \bm u_j, \bm u_t, \bm u_g). 
\end{equation}
\subsection{Reduced-Order model}
In the VLIP model, the center of pressure (CoP), denoted as $\bm c$, is defined as the weighted average position of the feet, given by $\bm c = \lambda_L\, \bm p_{F_L} + \lambda_R\, \bm p_{F_R}$, where $\lambda_i = u_{g_i,z} / (u_{g_L,z} + u_{g_R,z})$ for $i \in \{L,R\}$. In the Harpy full-dynamics model, which uses a point foot, $\bm c$ equals the stance foot position during the SS phase. The VLIP model without thrusters is underactuated, but the addition of thrusters makes the system fully actuated and enables trajectory tracking. For using the VLIP model in QP controller, we project the VLIP model on the sagittal plane. Equation of motion are derived assuming no slip $\left( \ddot{ \bm c} = 0 \right)$ as follows:
\begin{equation}
\begin{aligned}
    m \ddot{p}_{B,x} &= - mgsin\left(\alpha \right) + u_{t,x} -\lambda_{x}\\ 
    m \ddot{p}_{B,z} &= - mgcosQP \left(\alpha \right) + u_{t,z} + \lambda_{z}\\ 
\end{aligned}
\label{eq:model_vlip}
\end{equation}
where $m$ represents the mass of the VLIP model, which in this case is the total mass of the system, and $ [u_{t,x},u_{t,z} ]$ denotes the thruster forces about the CoM. The constraint force $[\lambda_{x},\lambda_{z}]$ is established to maintain the leg length $r$ equal to the leg conformation. $\alpha$ is the slope inclination.
By solving the zero moment point (ZMP) about the center of motion would result in the following equation.
\begin{equation}
\begin{gathered}
        \left( m \ddot{P}_{B,x} + \lambda_{x} \right)P_{B,z} = \left( m \ddot{P}_{B,z} - \lambda_{z} \right) \left(c_{x} - P_{B,x} \right)
\end{gathered}
\label{eq:model_vlip_ZMP}
\end{equation}
The linear pendulum model can be enforced by setting $P_{B,z} = z_0$ and $\ddot{P}_{B,z} = 0$. By setting LIP constraints, equation \eqref{eq:model_vlip_ZMP} will be become
\begin{equation}
\begin{gathered}
        \ddot{P}_{B,x} = \frac{\left(- \lambda_{z} \right) \left(c_{x} - P_{B,x} \right)}{m z_{0}} - \frac{\lambda_{x}}{m}
\end{gathered}
\label{eq:model_vlip_acc}
\end{equation}
\section{Swing controller}
The swing leg joint controller is designed to follow the desired foot position. The reference trajectory is generated by one dimensional fourth-order bezier polynomial. 
\begin{equation}
\begin{aligned}
    b_i(s) = \sum_{k=0}^{M} \beta_{ik} \frac{M!}{k!(M-k)!} s^k (1-s)^{M-k}
\end{aligned}
\label{eq:bezier_traj}
\end{equation}
where $s$ is the normalized gait timing variable varying from $0$ to $1$. $\beta$ represents the bezier parameters. The initial and final velocity of the gait is set to zero by setting $\beta_{1} = \beta{0}$ and $\beta^{i}_{M} = \beta^{i}_{M-1}$. This is done to avoid a large impact. Target joint angles $\bm q_t$ were found using inverse kinematics. Let $q = [\gamma_H, \phi_H, \phi_K]^{\top}$ represent the joint angles of the legs. Given the trajectory $\bm q_t$, the swing joint controller $\bm u_{sw}$ can be derived using a simple PID controller.
\begin{equation}
\begin{aligned}
    \bm u_{sw} = K_p \bm e + K_i \int_{0}^{t} \bm e(\tau) d\tau + K_d \frac{\partial \bm e(t)}{\partial t}
\end{aligned}
\label{eq:swing torque}
\end{equation}
\section{Stance controller}
\subsection{Quadratic Programming (QP)}
The objective of the stance controller is to maintain friction cone constraints and propel the body to track reference position and velocity.
we employ the non-linear dynamical equation \eqref{eq:model_vlip_acc} and subsequently linearize it to derive the state-space model.
\begin{equation}
   \dot{\bm x} = A\bm x + B\bm u
\end{equation}
\begin{equation}
\begin{aligned}
    A &= \frac{\partial \mathrm{F}(x,u)}{\partial x} = \begin{bmatrix}
    0 & 1 \\
    \frac{\lambda_{z,0}}{m z_0} & 0\\
    \end{bmatrix} \\
    B &= \frac{\partial \mathrm{F}(x,u)}{\partial u} = \begin{bmatrix}
     0 & 0 \\
     -\frac{1}{m} & \frac{P_{B,x,0} - c_{x,0}}{mz_{0}} \\
     \end{bmatrix}
\end{aligned}
\label{eq:state-space-model}
\end{equation}
where, $\bm x = [P_{B,x}, \dot{P}_{B,x}]^{\top}$ is the states and $\bm u = [\lambda_{x},\lambda_{z}]^{\top}$ is the input. we further discretize the dynamics, which will be used in the QP formulation. 
\begin{equation}
   \bm x_{K+1} = A_{d}\bm x_{K} + B_{d}\bm u_{K}
\end{equation}
here, $A_{d} = \mathbb{I} + At\Delta $ and $B_{d} = B \Delta t$. We proceed with eliminating the states from the decision variables by expressing them as a function of current state $x_{0}$ and control efforts
\begin{equation}
   \bm x = F\bm x_{0} + G\bm u
\end{equation}
where
\begin{equation}
\begin{aligned}
   \bm x &= \left[x^\top_{0}, x^\top_{1}, x^\top_{2}, \hdots x^\top_{N}\right]^\top \\
   \bm u &= \left[u^\top_{0}, u^\top_{1}, u^\top_{2}, \hdots u^\top_{N-1} \right]^\top
\end{aligned}
\end{equation}
\begin{gather}
   F = 
   \begin{bmatrix}
   \mathbb{I} \\
   A_d \\
   \vdots \\
   A^{N}_d
   \end{bmatrix}
   G = 
    \begin{bmatrix}
    0 &  &   &  & \\
    B_d & 0 &  &  & \\
    \vdots & & \ddots & & \\
    A^{N-1}_dB_d & A^{N-2}_dB_d & \hdots &A_d B_d & B_d\\
    \end{bmatrix}
\end{gather}
QP cost is tracking error with decision variable as $\bm u$.
\begin{equation}
   \min_{\bm u}~\bm u^\top P\bm u + c^\top \bm u
\label{eq:qp-formulation}
\end{equation}
subject to
\begin{equation}
\begin{aligned}
    A_{in}\bm u < B_{in} \\
\end{aligned}
\label{eq:qp-constraints}
\end{equation}
\begin{gather}
    A_{in} = 
\begin{bmatrix}
\mathbb{I} &  &  & \\
-\mathbb{I} & &  & \\
\vdots &  & \ddots & \mathbb{I} \\
0 & \hdots &  & -\mathbb{I}  
\end{bmatrix}     
    B_{in} = 
\begin{bmatrix}
\bm u_{u,0} \\
-\bm u_{l,0} \\
\vdots \\
\bm u_{u,n} \\
-\bm u_{l,n} \\
\end{bmatrix}     
\end{gather}
Here, $\bm u_{u}$ and $\bm u_{l}$ are the bounds from the friction cone. Friction cone bounds are $\lambda_{z} > \lambda_{min}$ and $\lambda_{x} < \mu |\lambda_{z}|$. We have constrained QP with inequality constraints and dense matrices, where $P$ and $c$ are defined as follows:
\begin{align*}
    P & = G^\top QG + R\\
    c & = \left(\bm x_{0}- \bm x_{d,0}\right) F^\top QG \\
\label{eq:qp-matrices}
\end{align*}
In this case, $\bm x_{0} $ are the initial states and $\bm x_{d}$ are the reference for those states. $Q$ is cost associated with error $\left(\bm x-\bm x_{d}\right)$ and $R$ is cost for control effort $\bm u = [\lambda_{x},\lambda_{z}]^{\top}$. 
\begin{figure}[t]
\centering
\includegraphics[width=1\linewidth]{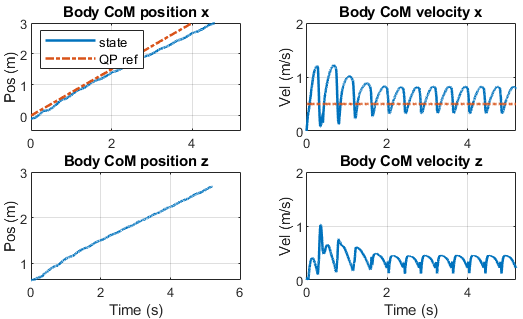}
    \caption{Shows the Harpy states and QP references throughout the simulation. The body's velocity states are stable during inclined walking.}
    \label{fig:states}
\end{figure}
\begin{figure}[t]
\centering
\includegraphics[width=1\linewidth]{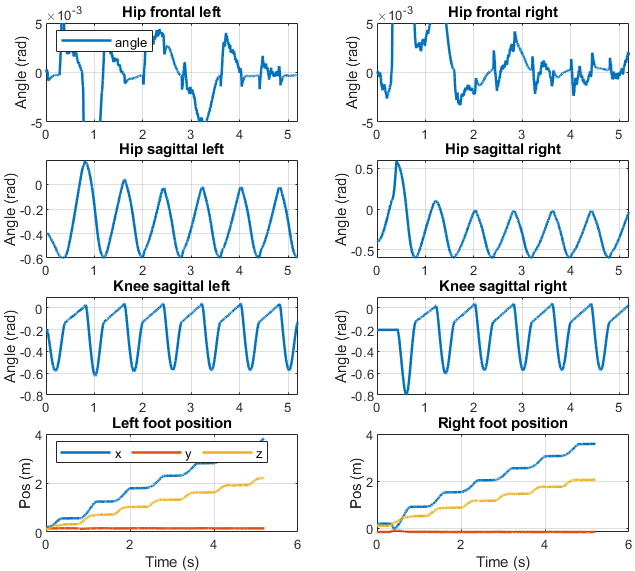}
    \caption{Illustrates Harpy's joint angle and foot positions in the simulation. Foot positions are defined in world frame.}
    \label{fig:joints}
\end{figure}
\subsection{Whole body control}
In previous section, QP solver outputs ground reaction forces $\left[\lambda_{x},\lambda_{z}\right]^\top$. Given ground reaction force and acceleration of the body, we used equation \eqref{eq:model_vlip} to compute the generalized thruster force $\bm u_{t} = [u_{t,x},0,u_{t,z}]$. To find the stance leg joint torque, we create a mapping matrix. we first simplify the model by considering massless linkages with concentrated mass at each joint and a mass and inertia at the body center. The states of the robot remain the same as that of the formulation shown in the~\eqref{eq:states}. The equation of motion is derived using Euler Lagrangian dynamics formulation. 
\begin{equation}
\begin{gathered}
    M \ddot{\bm q} + \bm h = B_j\, \bm u_j + \bm u_t + B_g\, \bm u_g
\end{gathered}
\label{eq:Mapping_eom}
\end{equation}
In this case, stance leg $\bm u_{g}$ is given by the Qp solver and swing leg $\bm u_g$ is equal to zero. Given that the QP solver ensures adherence to the friction cone condition, it is feasible to implement stance contact force constraints. 
\begin{equation}
\begin{gathered}
    J_s \ddot{\bm q} = - \dot{J}_s \dot{\bm q}
\end{gathered}
\label{eq:contact_constraint}
\end{equation}
where $J_s = [\partial \dot{\bm P}_F / \partial \dot{\bm q}]^{\top}$ is the contact jacobian. Further, we use $2$D reduced-order model and thus we implement planner constraint.
\begin{equation}
\begin{gathered}
    J_c \ddot{\bm q} = [0,0,0]^{\top}
\end{gathered}
\label{eq:planner constraint}
\end{equation}
here, $J_c$ is constraint jacobian and it ensures $y$, roll and yaw acceleration are set to zero. we combine equation \eqref{eq:swing torque}, \eqref{eq:Mapping_eom},\eqref{eq:contact_constraint} and \eqref{eq:planner constraint} to compute stance joint torque. 
\begin{equation}
\begin{gathered}
    \begin{bmatrix}
    M & -S_{st} & - J_{c}^{\top} \\
    J_s & 0_{3 \times 3} & 0_{3 \times 2} \\
    J_{c} & 0_{3 \times 3} & 0_{3 \times 2}
    \end{bmatrix} 
    \begin{bmatrix}
    \ddot{\bm q} \\
     \bm u_{st}\\
    \bm \lambda_{c}
    \end{bmatrix}
    = 
    \begin{bmatrix}
     - \bm h + S_{sw}\, \bm u_{sw} + \bm u_t + B_g\, \bm u_g \\
    -\dot{J}_s \dot{\bm q} \\
     [0,0,0]^{\top}
    \end{bmatrix} 
\end{gathered}
\label{eq:Mapping_matrix}
\end{equation}
here, $S_{sw}$ and $S_{st}$ are swing and stance selection matrix. $\lambda_c$ is the constraint force which ensures $y$ , roll and yaw acceleration remains zero. $u_{st}$ is the stance torque.
\section{results}
The numerical simulation (depicted in Fig. \ref{fig:snapshot}) and optimization are performed on Matlab where we use RK4 scheme to propagate the model forward in time. The Condensed Quadratic Programming (QP) formulation \eqref{eq:qp-formulation}, utilizes the qpSWIFT package \cite{pandala_qpswift_2019}. The Qp is running at $100$ Hz and numerical simulation is running at $2000$ Hz. 

\subsection{Simulation Specifications}
In this section, all units are in N, kg, m, s. The left leg of robot as following dimensions: $\bm l_1 = [0, 0.1, -0.1]^{\top}$, $\bm l_2 = [0, 0.5, 0]^{\top}$, $\bm l_3 = [0, 0, -0.3]^{\top}$ and $\bm l_4 = [0, 0.1, 0]^{\top}$. The right side has the $y$ axis component inverted. The following mass and inertias were used for simulation: $m_{B} = 4$, $m_{H} =m_{K} = 0.5$, $I_{B} = 10^{-3}$ and $I_{H} = I_{K} = 10^{-4}$. For whole body controller, we only consider mass and body inertia. Finally, ground parameters were $\mu_s = 0.8$, $\mu_c = 0.64$, $\mu_v = 0.8$, $k_{g,p} = 8000$ and $k_{g,d} = 268$. 
The weighting matrices for QP were $Q = diag([300000, 2000])$ and $R = diag([1, 1])$. Slope inclination $\alpha$ is $30^{\circ}$.
\begin{figure}[t]
\centering
\includegraphics[width=1\linewidth]{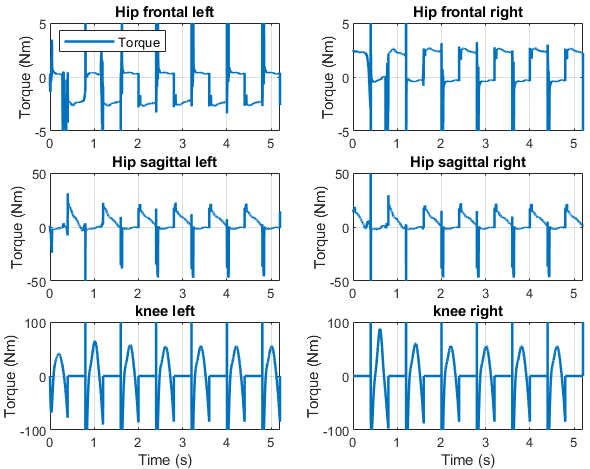}
    \caption{Illustrates simulated joint torques during the inclined walking.}
    \label{fig:torque}
\end{figure}
\begin{figure}[t]
\centering
\includegraphics[width=1\linewidth]{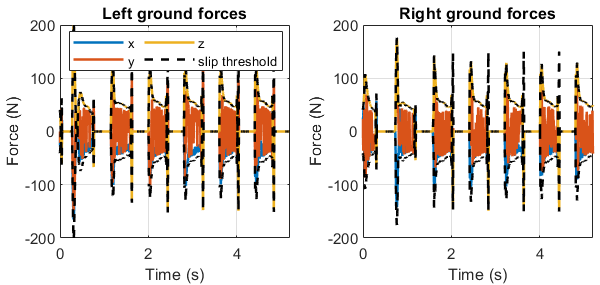}
    \caption{Illustrates ground reaction forces from the ground model in the simulation.}
    \label{fig:GRF}
\end{figure}
\subsection{Results and Discussion}
\begin{figure}[t]
\centering
\includegraphics[width=0.9\linewidth]{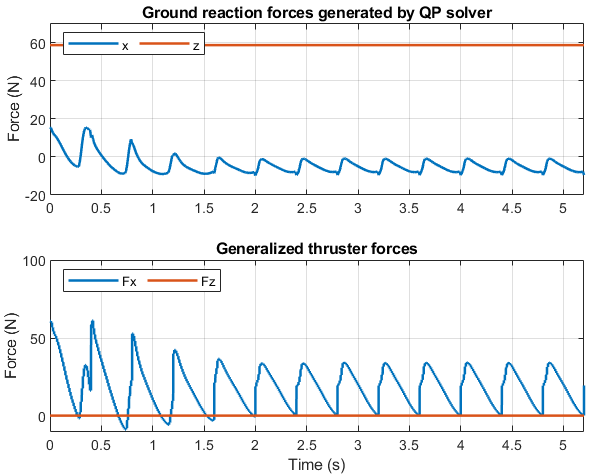}
    \caption{Illustrates $\lambda_{x}$,  $\lambda_{z}$ generated by QP solver and calculated thruster forces.}
    \label{fig:QP-output}
\end{figure}

Fig.~\ref{fig:states} illustrates the robot’s position and velocity along the $x$ and $z$ axes. It is evident that the robot closely adheres to the reference trajectory, which is the primary objective of the Quadratic Programming (QP) approach. Fig.~\ref{fig:QP-output} depicts the Ground Reaction Forces (GRF) generated by the QP solver, consistently maintaining compliance with the friction cone condition, thereby preventing slippage. The robot uses a higher thruster force initially to attain the desired velocity. Fig.~\ref{fig:torque} presents the torque computed using the GRF from the QP solver for the stance phase, and the reference trajectory for the swing leg. Fig.~\ref{fig:GRF} shows the ground reaction force generated by the complaint ground model while the robot is walking on an inclined slope. dotted line shows the friction cone threshold and it can be seen that even when Qp optimization is not running robot never violates friction cone condition. Finally, Fig.~\ref{fig:joints} displays the joint angles and foot positions of the left and right legs in the world frame.

\section{conclusion}
In this paper, we propose a QP formulation that is computationally efficient, and algorithmically robust. The proposed algorithm ensures internal and contact constraints are satisfied. It uses QP and a reduced-order model to compute reaction forces satisfying contact constraints and then provides the thruster forces. The whole body controller then maps these forces on a high-fidelity model to produce thruster-assisted slope walking. In our numerical simulation tests, QP computes contact force forces for a robot in 0.3ms. The swing leg controller provides the required torque to track the generated open-loop trajectory and complete the gait cycles. The QP solver was quickly able to achieve a stable limit cycle showing the efficiency of the control method.
Future work will focus on extending the method to 3D motion. Also, the current formulation uses a PID controller for the swing leg, which can be replaced by a heuristic method such as Raibert's control or capture point control. 










\printbibliography

\end{document}